\documentclass[conference]{IEEEtran}
\IEEEoverridecommandlockouts
\usepackage{cite}
\usepackage{amsmath,amssymb,amsfonts}
\usepackage{algorithmic}
\usepackage{graphicx}
\usepackage{textcomp}
\usepackage{xcolor}
\usepackage{booktabs}
\usepackage{pgf-pie}
\usepackage{url}
\usepackage{hyperref}
\usepackage{tikz}
\usetikzlibrary{arrows.meta, positioning, calc}
\usepackage{pgfplots}
\pgfplotsset{compat=1.18}

\def\BibTeX{{\rm B\kern-.05em{\sc i\kern-.025em b}\kern-.08em
    T\kern-.1667em\lower.7ex\hbox{E}\kern-.125emX}}

\begin{document}




\title{An Expanded Synthetic Conversation Dataset\\
for Multi-Turn Smishing Detection}
\author{
\IEEEauthorblockN{Carl Lochstampfor}
\IEEEauthorblockA{\textit{Department of Computer Science}\\
\textit{Old Dominion University}\\
Norfolk, VA, USA\\
cloch001@odu.edu}
\and
\IEEEauthorblockN{Ayan Roy}
\IEEEauthorblockA{\textit{Department of Computer Science}\\
\textit{Christopher Newport University}\\
Newport News, VA, USA\\
ayan.roy@cnu.edu}
}

\maketitle

\begin{abstract}
Our prior work introduced COVA, a synthetically generated multi-turn
conversational smishing dataset of 3,201 labeled conversations, establishing
baseline detection benchmarks across eight models. While XGBoost with TF-IDF
features achieved the best performance (72.5\% accuracy, 0.691 macro F1),
transformer models underperformed---attributed to input truncation and
insufficient training data. We present COVA-X, an expanded dataset of 10,985
conversations spanning eight elder-targeted scam categories, produced by an
improved generation pipeline addressing contamination, label mismatch,
stage-direction bleed, and prompt-design failures from the first iteration.
Retraining all classifiers on the expanded dataset yields the central finding
of this work: \textbf{Longformer now surpasses XGBoost on all evaluation
metrics} (79.71\% accuracy, 0.7786 macro F1 vs.\ 78.43\% and 0.7563),
directly confirming that transformer models require larger conversational
corpora to realize their contextual advantages. We additionally document a
quality lifecycle including a 12.7$\times$ improvement in label correction
rate (3.9\% vs.\ 49.8\%), an architectural intervention reducing
virtual-kidnapping artifact rates (67.1\%$\rightarrow$46.5\%), and a
per-scam-type outcome analysis showing scam categories modulate results in
mechanism-consistent ways. A pre/post-cleanup sensitivity analysis confirms
that dataset refinement recovers genuine label-relevant signal across all
three classifier architectures.
\end{abstract}

\section{Introduction}

Conversational social engineering attacks targeting elderly individuals
represent a growing and underserved threat in the cybersecurity landscape.
Unlike single-message phishing, these attacks unfold across multiple
dialogue turns, gradually building trust before extracting financial or
personal information~\cite{b1,b2}. Real-world data on such interactions
is scarce due to privacy and ethical constraints, motivating the use of
synthetic data generation.

Earlier we introduced COVA ~\cite{cova1} , the first publicly available
multi-turn conversational smishing dataset, comprising 3,201 labeled
conversations across eight elder-targeted scam categories. That work
established eight baseline classifiers and identified a critical limitation:
despite their superior contextual language understanding, transformer models
(DistilBERT and Longformer) underperformed compared to XGBoost with TF-IDF
features. We attributed this gap to two factors: (1) input truncation
discarding outcome-critical final turns in DistilBERT, and (2) insufficient
training data for effective transformer fine-tuning. We explicitly predicted
that ``transformer models would likely benefit from larger training corpora
($>$5,000 conversations).''

This paper presents COVA-X, the expansion of the COVA dataset to 10,985
conversations (3.43$\times$ increase),\footnote{Per-scam-type generation
targets summed to approximately 11,400; the realized production count of
10,985 reflects normal generator retry overhead and minimum-turn-floor
discards. Comprehensive dataset documentation, including version
history, per-scam-type counts, and known limitations, is provided in
the COVA-X data sheet, released as a companion artifact alongside the dataset.} and reports
the results of retraining all classifiers on the expanded dataset. The
central finding confirms the first paper's hypothesis:
\textbf{Longformer now surpasses XGBoost on all metrics for the first
time}, achieving 79.71\% accuracy and 0.7786 macro F1 on the 3-class
outcome prediction task.
Beyond model performance improvements, this paper presents a comprehensive dataset-quality lifecycle analysis. This includes pipeline failures identified and corrected during dataset expansion, a three-point evidence chain evaluating an architectural intervention for reducing virtual-kidnapping artifacts, improvements in label correction rates, the quantitative contribution of each pipeline stage to dataset integrity, and a pre/post-cleanup classifier sensitivity analysis demonstrating that the cleanup improved label-relevant signal across three classifier architectures. We argue that transparent documentation of these iterative challenges constitutes a methodological contribution in its own right, offering a practical roadmap for future synthetic dataset construction efforts.

The main contributions of this paper are:
\begin{itemize}
    \item We expand the COVA dataset from 3,201 to 10,985
    conversations across eight elder-targeted scam categories, with
    five victim and attacker profiles per type.
    \item We document and quantify a full quality lifecycle pipeline
    including contamination scanning, stage-direction stripping,
    automated relabeling, and a three-role architectural intervention
    for virtual-kidnapping generation, achieving a 3.9\% label
    correction rate versus 49.8\% in the first iteration and a
    reduction in virtual-kidnapping artifact rates
    (67.1\%$\rightarrow$46.5\%).
    \item We report a per-scam-type outcome distribution analysis
    showing that scam-type characteristics modulate outcomes in
    mechanism-consistent ways: virtual kidnapping has the highest
    \texttt{successful\_scam} rate (33\%) due to emotional-pressure
    framing; grandparent has the highest \texttt{verification\_attempt}
    rate (63\%) consistent with verification-call-back scripts; and
    medicare/bank have substantial \texttt{quick\_rejection}
    populations reflecting public awareness.
    \item We retrain all classifiers and confirm that Longformer
    surpasses XGBoost on the expanded dataset on every reported
    metric, validating the data-size hypothesis from the first paper.
    A pre/post-cleanup sensitivity analysis shows all three
    architectures improve on the cleaned dataset, evidence the cleanup
    affected genuine label-relevant signal.
    \item We characterize Qwen 2.5 14B generation behavior under
    sustained emotional pressure, documenting a five-instance pattern
    of explicit prompt-directive override across distinct directive
    types (structural-tag adoption, victim-profile compliance,
    address-term parameterization, name-identity rules, and
    conversation-termination directives) that motivates the
    model-comparison study proposed in future work.
\end{itemize}

\section{Related Work}
\label{sec:related}

\subsection{Social Engineering Detection}

Detection of social engineering attacks has been studied across multiple
modalities. In the SMS domain, Seo et al.~\cite{b3} proposed on-device
smishing classifiers resistant to text evasion, while Patra et al.~\cite{b4}
developed prediction models combining machine learning with text analysis.
These approaches address single-message classification and do not capture
multi-turn dynamics.

Derakhshan et al.~\cite{b5} introduced ASsET, a detection system based on
scam signatures for telephone-based attacks. Wood et al.~\cite{b6} analyzed
scam-baiting calls from YouTube, identifying scam stages and scripts at
scale. Jain et al.~\cite{b7} demonstrated that BERT outperforms TF-IDF for
single-message smishing detection, a result that motivates investigating
whether this advantage transfers to conversational detection with
sufficient data.

Recent work has extended detection beyond conventional text classification.
Park et al.~\cite{b15} present an LLM-based framework for voice-phishing
detection that combines case-informed transcript generation with
domain-expert prompt engineering, achieving substantially improved
classification accuracy on Korean voice-phishing data through synthetic
augmentation. Their work establishes synthetic transcript generation as a
viable strategy for addressing phishing data scarcity. Closely related,
Li et al.~\cite{b14} examine the adversarial dimension of LLM-generated
phishing content, demonstrating that LLMs can be used by attackers to
construct adversarial transcripts capable of evading ML-based detectors.
Their results inform our own threat-model framing: the same generation
capability that enables defensive dataset construction also enables
adversarial transcript synthesis. Gressel et al.~\cite{b16} address the
romance-scam category specifically, examining LLMs' role in detecting
emotionally manipulative scam dialogue, a category with particular
relevance to our COVA-X romance subset (the largest single category at
1,934 conversations). For real-time deployment, the AI-in-the-Loop
framework~\cite{b17} proposes federated and privacy-preserving
classification of scam dialogue from partial transcripts, which informs
the incremental-classification direction proposed in
Section~\ref{sec:future_work}.

\subsection{Synthetic Data and Multi-Agent Simulation}

Basta et al.~\cite{b8} presented ``Bot Wars,'' a framework using competing
LLMs as scam-baiters, validated against 179 hours of human interactions.
Kumarage et al.~\cite{b9} proposed SE-VSim, an LLM-agentic framework for
simulating social engineering in multi-turn conversations with varying
victim personality traits. Spokoyny et al.~\cite{b10} developed CHATTERBOX
for automated long-term engagement with online scammers. These frameworks
share with our work a multi-agent design, but differ in deployment posture:
prior work focuses on adversarial engagement against active scammers,
whereas COVA-X uses multi-agent simulation as a dataset-construction
methodology for downstream classifier training.

A practical concern relevant to our prompt-engineering methodology is
what we term a salience-bias effect in instruction-tuned LLMs:
enumerating prohibited patterns as negative examples can increase
rather than decrease their production rate, a phenomenon documented
across both general instruction-following~\cite{b18, b19} and
negation-specific generation contexts~\cite{b20}.
We observe this effect directly in our virtual-kidnapping
attacker prompt iterations (Section~\ref{sec:discussion}, Salience Bias
in Prompt Engineering subsection).

\subsection{Gap in the Literature}

The COVA-X expansion addresses gaps identified in our prior work and the
broader literature: (1) limited dataset scale constraining transformer
evaluation; (2) single victim profiles limiting behavioral diversity;
(3) undocumented pipeline quality lifecycles making reproducibility
difficult; (4) limited characterization of generation-model capability
limits encountered during synthetic dataset construction. Herrera et
al.~\cite{b11} highlight that older adults remain disproportionately
vulnerable to AI-enhanced scams, reinforcing the importance of
large-scale, multi-category elder fraud datasets.

\section{Dataset Expansion Methodology}
\label{sec:methodology}

\subsection{Generation Infrastructure}

The COVA-X expansion used the same core framework as the first dataset:
two independent LLM agents (attacker and victim) exchanging turns using
Qwen 2.5 14B via Ollama on local GPU hardware, with a three-role
extension for virtual kidnapping (Section~\ref{sec:discussion}). All
generation uses local inference rather than commercial API access, for
three reasons: research freedom on simulated scam dialogue, zero
marginal cost for bulk generation, and reproducibility through
Ollama-pinned model weights.

Expansion runs were distributed across two workstations:

\begin{itemize}
    \item \textbf{Workstation A}: NVIDIA RTX 4080 Super 16GB GDDR6X ---
    primary generation and classifier-training machine; used for romance,
    government impersonation, medicare, and virtual kidnapping.
    \item \textbf{Workstation B}: NVIDIA RTX 5060 Ti 16GB GDDR7 (Blackwell
    architecture) --- secondary generation machine; used for grandparent,
    bank, lottery, and investment.
\end{itemize}

Both workstations ran Qwen 2.5 14B at 16k context. Sampling parameters
were temperature 0.8 for attacker and victim turns, and temperature 0.7
with a 60-token cap for virtual-kidnapping three-role hostage utterances
(the token cap enforces the 3--12 word brevity constraint for hostage
voice). These settings produce stochastic variation across regeneration
runs even with identical seeds; reproducibility of exact conversation
content is therefore not guaranteed, but distributional characteristics
(per-scam-type counts, outcome distribution, average turn lengths,
quality-flag rates) are reproducible within sampling-noise margins.
Conversations under 10 turns were discarded as insufficiently developed
(skip rate approximately 2\% overall, 25--30\% for virtual kidnapping).

\subsection{Profile Expansion}

Cova \cite{cova1} used a single victim profile per scam type. COVA-X
introduces five profiles per type: two outcome-pinned profiles (Carol
pinned to \texttt{quick\_rejection}, George pinned to
\texttt{scam\_detected}) and three general-pool profiles covering the
remaining outcomes. This design provides sufficient behavioral variety
while maintaining enough conversations per profile for classifier
signal strength. Profile-pinning is implemented through outcome-target
dictionaries in the generator, with hard-stop directives written into
the system prompt for the pinned outcome targets.

\subsection{Dataset Targets and Distribution}

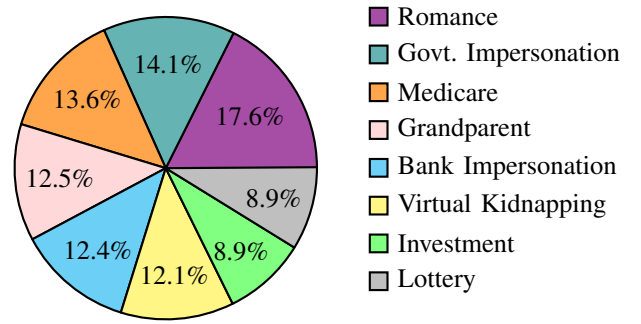
\begin{figure}[h]
\centering
\begin{tikzpicture}
  \pie[
    radius=2.0,
    color={
      violet!70,
      teal!60,
      orange!70,
      pink!60,
      cyan!50,
      yellow!60,
      green!50,
      gray!50
    },
    text=legend,
    before number=,
    after number=\%,
    sum=100
  ]{
    17.6/Romance,
    14.1/Govt.\ Impersonation,
    13.6/Medicare,
    12.5/Grandparent,
    12.4/Bank Impersonation,
    12.1/Virtual Kidnapping,
     8.9/Investment,
     8.9/Lottery
  }
\end{tikzpicture}
\caption{COVA-X per-scam-type distribution ($n=10{,}985$).}
\label{fig:scam_pie}
\end{figure}

Per-scam-type generation targets summed to approximately 11,400
conversations across the eight scam categories. The realized
production output, after normal generator retry overhead and
minimum-turn-floor discards, is 10,985 conversations.
Fig~\ref{fig:scam_pie} shows the per-scam-type production counts;
the COVA-X data sheet documents the per-scam-type generation
targets and the production-vs-target reconciliation in detail.

\subsection{Prompt Engineering Improvements}

Several prompt-level fixes were introduced for the expansion based on
failures observed in the first dataset:

\begin{itemize}
    \item \textbf{Name Identity Lock}: A zero-tolerance name rule added
    to all attacker and victim prompts, preventing character name mixing
    across scam types.
    \item \textbf{ABSOLUTE OUTPUT RULE}: Injected at runtime into victim
    prompts to suppress stage-direction bleed. Most effective for
    low-emotional-pressure scam types; partially effective for virtual
    kidnapping and romance, where additional architectural and
    post-processing measures were needed (Section~\ref{sec:quality}).
    \item \textbf{Grandparent-Opener Contamination Fix}: Attacker prompts
    for government impersonation, medicare, and virtual kidnapping were
    rewritten to eliminate ``Hi Grandma/Grandpa'' openers that had
    infected all non-grandparent types in ~\cite{cova1}. Residual
    legacy contamination from pre-fix runs is documented in
    Section~\ref{sec:quality}.
    \item \textbf{Pipeline Bug Fix---Prompt Loading}: A hardcoded
    \texttt{\_v1.md} suffix in \texttt{load\_prompt\_template()} silently
    prevented updated prompts from loading in the first dataset. Fixed
    via glob-pattern matching for all prompt versions.
\end{itemize}

\subsection{Per-Scam-Type Outcome Distribution}
\label{sec:outcome_distribution}

A property of the COVA-X dataset that emerges from the production-count
review is the per-scam-type outcome distribution. Rather than appearing
as a roughly uniform pattern across types, outcomes are strongly
modulated by scam-type characteristics: scams that exploit emotional
pressure (virtual kidnapping) produce different outcome distributions
than scams that exploit institutional trust (medicare, bank), which
themselves differ from scams that exploit gradual relationship
formation (romance). Table~\ref{tab:outcome_distribution} reports the
per-scam-type, per-outcome counts.

\begin{table}[]
\caption{COVA-X outcome distribution by scam type across five terminal states: \textit{succ.}, \textit{part.}, \textit{ver.}, \textit{det.}, and \textit{rej.}}
\label{tab:outcome_distribution}
\centering
\footnotesize
\begin{tabular}{lrrrrr}
\toprule
\textbf{Scam Type} & \textbf{succ.} & \textbf{part.} & \textbf{ver.} & \textbf{det.} & \textbf{rej.} \\
\midrule
Romance              & 254 & 921 & 481 & 267 & 11 \\
Government Imp.      & 202 & 468 & 551 & 294 & 30 \\
Medicare             & 254 & 376 & 268 & 159 & 433 \\
Virtual Kidnapping   & 443 & 659 & 110 & 107 & 11 \\
Grandparent          &  63 & 296 & 868 & 144 &  5 \\
Bank Impersonation   & 173 & 377 & 569 &  29 & 212 \\
Lottery              & 196 & 477 & 290 &  10 &  2 \\
Investment           &  86 & 705 & 146 &   5 & 33 \\
\bottomrule
\end{tabular}
\end{table}

Three patterns are worth noting because they shape the classifier
analysis in Section~\ref{sec:experimentation}. First, virtual
kidnapping has the highest \texttt{successful\_scam} rate of any type
(443 of 1,330, or 33\%), more than double the dataset-wide
\texttt{successful\_scam} rate (1,671 of 10,985, or 15.2\%). This is
consistent with the high-emotional-pressure framing exploited by VK
scams: simulated victims under acute stress are markedly more likely
to comply with attacker demands. Second, grandparent has by far the
highest \texttt{verification\_attempt} rate (868 of 1,376, or 63\%),
consistent with the grandparent-scam pattern in which victims attempt
to verify the call back to the supposed grandchild before acting.
Third, medicare and bank scams have substantial \texttt{quick\_rejection}
populations (433 and 212 respectively), reflecting the public salience
of medicare and bank scams and victims having institutional alternatives
(call the bank directly, hang up).

These patterns motivate the per-scam-type breakdown reported alongside
classifier results in Section~\ref{sec:experimentation}: a classifier's
overall accuracy is shaped by the dataset's outcome distribution, and
per-type breakdowns reveal architecture-specific strengths obscured by
aggregate metrics. The outcome-distribution heterogeneity 
motivates the stratified train/val/test split (scam type $\times$
outcome class, seed 42), preserving the joint distribution of these
two variables across splits.

\section{Dataset Quality Lifecycle}
\label{sec:quality}

\subsection{Post-Generation Quality Scan}

All 10,985 preprocessed conversations were scanned using
\texttt{scan\_generation\_quality.py}, which detects five artifact
types: DIALOGUE\_COLLAPSE (victim writing attacker lines in their own
turn), STAGE\_DIRECTION (bracket/parenthetical stage directions),
TURN\_OVERFLOW (excessively long individual turns), META\_BLEED (model
instruction echo), and LOOP\_DETECTION (repetitive content).

Across the full preprocessed dataset, the post-cleanup flag rate is
14.5\%, dominated by virtual-kidnapping conversations whose
post-cleanup flag rate is 46.5\% (Section~\ref{sec:vk_analysis}).
Without virtual kidnapping, the flag rate across all other types is
approximately 3.5\%. Table~\ref{tab:per_type_artifacts} reports the
per-scam-type breakdown of post-cleanup flag rates and shows the
heterogeneity that motivated the architectural intervention reported
in Section~\ref{sec:vk_analysis}.

\begin{table}[h]
\caption{Post-Cleanup Flag Rates by Scam Type. Virtual kidnapping is
the dominant source of artifact concentration; remaining types are
all below 6\%.}
\label{tab:per_type_artifacts}
\centering
\footnotesize
\begin{tabular}{lrrr}
\toprule
\textbf{Scam Type} & \textbf{Flagged} & \textbf{Total} & \textbf{Rate} \\
\midrule
Virtual Kidnapping       & 619  & 1,330  & 46.5\% \\
Romance                  & 78   & 1,934  & 4.0\%  \\
Investment               & 47   & 975    & 4.8\%  \\
Lottery                  & 36   & 975    & 3.7\%  \\
Bank Impersonation       & 47   & 1,360  & 3.5\%  \\
Grandparent              & 38   & 1,376  & 2.8\%  \\
Government Impersonation & 41   & 1,545  & 2.7\%  \\
Medicare                 & 87   & 1,490  & 5.8\%  \\
\midrule
\textbf{Total}           & \textbf{993} & \textbf{10,985} & \textbf{9.0\%} \\
\textbf{Excl.\ VK}       & \textbf{374} & \textbf{9,655}  & \textbf{3.9\%} \\
\bottomrule
\end{tabular}
\end{table}

The dataset-wide rate of 14.5\% reported in earlier analysis includes
multiple flag types per conversation; Table~\ref{tab:per_type_artifacts}
reports unique conversations flagged at least once. Both framings
agree on the central observation: VK is the dominant driver of
artifact concentration, and remaining scam types are individually
below 6\%.

\subsection{Virtual-Kidnapping Three-Point Evidence Chain}
\label{sec:vk_analysis}

Virtual kidnapping is uniquely structured among the eight scam types:
it involves three parties (kidnapper, victim, and a supposedly-kidnapped
hostage). Initial generation used a two-role architecture, in which the
attacker model was directed to produce hostage utterances within its own
turns using a structural \texttt{[HOSTAGE VOICE]:} tag. This produced a
67.1\% flag rate on virtual-kidnapping conversations, dominated by
DIALOGUE\_COLLAPSE (the attacker model writing all three voices within
a single turn) and STAGE\_DIRECTION leakage.

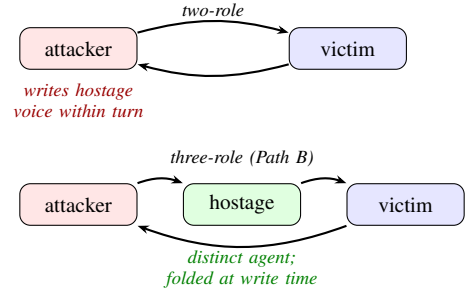
\begin{figure}[h]
\centering
\begin{tikzpicture}[
    node distance=0.65cm and 0.65cm,
    role/.style={rectangle, draw, rounded corners, minimum width=1.55cm,
                 minimum height=0.55cm, font=\footnotesize, align=center},
    arr/.style={-{Stealth[length=1.5mm]}, thick},
    label/.style={font=\scriptsize\itshape, align=center}
]
\node[role, fill=red!10] (a1)  {attacker};
\node[role, fill=blue!10, right=2.0cm of a1] (v1) {victim};
\node[label, below=0.05cm of a1, text=red!60!black]
      {writes hostage\\voice within turn};
\draw[arr, bend left=15] (a1) to (v1);
\draw[arr, bend left=15] (v1) to (a1);
\node[label, above=0.3cm of $(a1)!0.5!(v1)$] { two-role};

\node[role, fill=red!10, below=1.5cm of a1] (a2)  {attacker};
\node[role, fill=green!12, right=0.6cm of a2] (h2) {hostage};
\node[role, fill=blue!10, right=0.6cm of h2] (v2) {victim};
\node[label, below=0.2cm of h2, text=green!50!black]
      { distinct agent;\\folded at write time};
\draw[arr, bend left=20] (a2.north east) to (h2.north west);
\draw[arr, bend left=20] (h2.north east) to (v2.north west);
\draw[arr, bend left=20] (v2.south west) to (a2.south east);
\node[label, above=0.1cm of h2] {three-role (Path B)};
\end{tikzpicture}
\caption{Generation architectures for virtual kidnapping. In the two-role design, the attacker model generates hostage speech within its own turns, increasing the risk of dialogue collapse. In three-role Path B, the hostage is modeled as a separate agent with its own system prompt; hostage turns are later merged into the preceding attacker turn using a \texttt{[HOSTAGE VOICE]:} tag.}
\label{fig:architecture}
\end{figure}

Section~\ref{sec:discussion} (Architectural Workaround and Its Residual
Limits) documents the three-role generation architecture (Path B) we
introduced to address this. Path B treats the hostage as a distinct
agent during generation, with a dedicated system prompt and its own
turn; hostage turns are folded backward into the preceding attacker
turn at JSON-write time with the \texttt{[HOSTAGE VOICE]:} tag added by
the generator itself. Figure~\ref{fig:architecture} contrasts the
two architectures.

Table~\ref{tab:vk_chain} reports the three-point evidence chain
documenting Path B's effect on virtual-kidnapping flag rates: the
pre-Path-B baseline, the post-Path-B raw rate (after architectural
intervention but before post-processing), and the post-Path-B
processed rate (after a bare-colon stripping filter and a
non-Latin-script filter applied during cleaning).

\begin{table}[h]
\caption{Three-Point Evidence Chain: VK Flag Rate by Pipeline Stage}
\label{tab:vk_chain}
\centering
\begin{tabular}{lr}
\toprule
\textbf{Pipeline Stage} & \textbf{VK Flag Rate} \\
\midrule
Pre-Path-B (two-role baseline)             & 67.1\% \\
Post-Path-B raw (three-role, no cleanup)   & 46.7\% \\
Post-Path-B processed (after cleaning)     & 46.5\% \\
\bottomrule
\end{tabular}
\end{table}

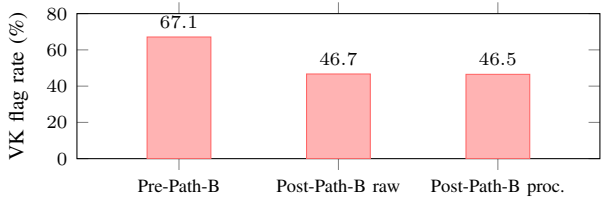
\begin{figure}[]
\centering
\begin{tikzpicture}
\begin{axis}[
    width=8.5cm,
    height=3.5cm,
    ybar,
    bar width=24pt,
    enlarge x limits=0.32,
    ymin=0, ymax=80,
    ylabel={VK flag rate (\%)},
    symbolic x coords={Pre-Path-B, Post-Path-B raw, Post-Path-B proc.},
    xtick=data,
    nodes near coords,
    nodes near coords align={vertical},
    label style={font=\footnotesize},
    tick label style={font=\scriptsize},
    every node near coord/.append style={font=\scriptsize}
]
\addplot[fill=red!30, draw=red!60] coordinates {
    (Pre-Path-B, 67.1)
    (Post-Path-B raw, 46.7)
    (Post-Path-B proc., 46.5)
};
\end{axis}
\end{tikzpicture}
\caption{Virtual-kidnapping flag rates across pipeline stages: the three-role intervention reduced flags by 20.4 points, with post-processing reducing them by only 0.2 more, indicating that most removable artifacts were eliminated by construction rather than downstream cleanup.}
\label{fig:vk_trajectory}
\end{figure}

Figure~\ref{fig:vk_trajectory} visualizes the trajectory across these
three pipeline points. The visual representation makes vivid that the
architectural intervention accounts for the bulk of the artifact-rate
reduction (20.4 \% points), while post-processing contributes
a much smaller incremental effect (0.2\% points).

Majority of the reduction (20.4 \% points) comes from the
architectural change. The post-processing stage contributes a smaller
0.2\% point further reduction, indicating that most cleanable
artifact patterns were eliminated by construction once the three-role
separation was in place. The residual 46.5\% flag rate reflects
artifact categories that the architectural intervention does not
address (repetition loops under emotional pressure, address-term
distribution bias) and is documented as a Qwen 2.5 14B capability
limit in Section~\ref{sec:discussion}.

\subsection{Contamination Scan}

Scanning for grandparent-opener bleed and cross-scam contamination
revealed that the contamination was successfully eliminated for bank,
lottery, and investment (0 grandparent openers). Government
impersonation (357 conversations), medicare (348), and virtual
kidnapping (355) retain grandparent-opener contamination from pre-fix
pipeline runs performed during the initial expansion. These are documented as
legacy contamination rather than corrected post-hoc, preserving data
integrity and providing a measurable comparison point for future
pipeline iterations.

\subsection{Stage-Direction Stripping}

A post-processing pass using \texttt{strip\_stage\_directions()} was
applied to all preprocessed conversations. The pass affected
approximately 0.6\% of dataset turns (concentrated in virtual
kidnapping) with minimal modification to other scam types. This
minimal impact on non-VK types confirms that the ABSOLUTE OUTPUT RULE
injection during generation was largely effective for six of the
eight scam types; the residual VK rate is the focus of the
three-role architectural intervention reported above.

\subsection{Label Quality Audit and Relabeling}

The full dataset was audited using \texttt{audit\_outcome\_labels\_v2.py}
and relabeled via \texttt{relabel\_outcomes\_v2.py}. Of the
10,985 conversations processed by the audit, 439
(3.9\%) received corrected labels. Per-scam-type breakdown of the
439 corrections: medicare 195, bank 95, investment 50, virtual
kidnapping 32, romance 18, government impersonation 17, grandparent
16, lottery 16. The most common single correction was
\texttt{verification\_attempt $\rightarrow$ quick\_rejection} (121
instances). A second-pass audit using updated v2 heuristics
identified seven high-confidence Carol-profile compound-response
conversations that had been mislabeled as
\texttt{verification\_attempt}; these were corrected to
\texttt{quick\_rejection}. An additional 5,734 v2-audit
findings reflecting the v2 audit's expanded detection rules were
deferred to future work pending manual classification of true
mismatches versus rule false positives.

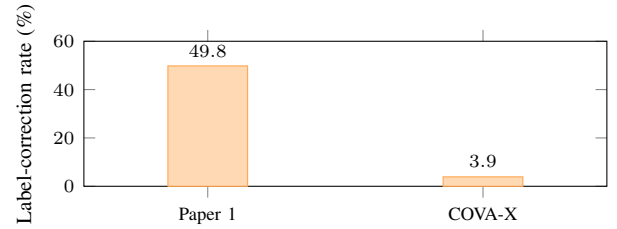
\begin{figure}[h]
\centering
\begin{tikzpicture}
\begin{axis}[
    width=8.5cm,
    height=3.5cm,
    ybar,
    bar width=30pt,
    enlarge x limits=0.45,
    ymin=0, ymax=60,
    ylabel={Label-correction rate (\%)},
    symbolic x coords={Paper 1, COVA-X},
    xtick=data,
    nodes near coords,
    nodes near coords align={vertical},
    label style={font=\footnotesize},
    tick label style={font=\scriptsize},
    every node near coord/.append style={font=\scriptsize}
]
\addplot[fill=orange!30, draw=orange!70] coordinates {
    (Paper 1, 49.8)
    (COVA-X, 3.9)
};
\end{axis}
\end{tikzpicture}
\caption{Label-correction rates fell from 49.8\% in Paper 1 to 3.9\% in COVA-X, a $12.7\times$ improvement. This reduction reflects three pipeline changes: outcome-pinned victim profiles, runtime ABSOLUTE OUTPUT RULE injection, and an expanded attacker-victim profile set that produced more self-consistent conversation arcs.}
\label{fig:label_correction}
\end{figure}

This represents a dramatic improvement over the first dataset's 49.8\%
mismatch rate---a 12.7$\times$ improvement in label self-consistency.
The improvement is attributable to three pipeline changes: better
prompt behavioral guidance, outcome-pinned profiles for Carol and
George, and the expanded attacker/victim profile set producing more
self-consistent conversations. Figure~\ref{fig:label_correction}
visualizes the comparison.

A pipeline bug was also discovered and fixed during this phase:
\texttt{relabel\_outcomes\_v2.py} used \texttt{glob()} rather than
\texttt{rglob()}, silently skipping all subdirectory files. The fix
was applied before the relabeling run.

\section{Experimentation}
\label{sec:experimentation}

\subsection{Experimental Setup}

\subsubsection{Data Splits}
The 10,985-conversation dataset\footnote{The COVA-X dataset is fully
documented in a companion data sheet covering identity, composition,
provenance, classifier results, and known limitations. The data sheet
will be released alongside the dataset.
Per-scam-type generation targets summed to approximately 11,400; the
realized production count of 10,985 reflects normal generator retry
overhead and minimum-turn-floor discards rather than separate
preprocessing-filter loss.} was partitioned into stratified train (80\%),
validation (10\%), and test (10\%) splits using random seed 42, maintaining
proportional representation of scam categories and outcome classes. This
yielded 8,787 training conversations, 1,099 validation conversations, and
1,099 test conversations. Total turn count after preprocessing is 88,852.

\subsubsection{Label Mapping}
The same 3-class collapse from the first paper was applied:
\textit{complied} (successful\_scam), \textit{partial}
(partial\_compliance), and \textit{rejected} (verification\_attempt +
scam\_detected + quick\_rejection). Class distribution in the training
set: rejected 46.4\%, partial 37.5\%, complied 16.1\%.

\subsubsection{Models}
We retrained the same model families as the first paper, focusing on
the three best-performing configurations: XGBoost + TF-IDF (sklearn),
DistilBERT v2 (tail truncation, 512 tokens), and Longformer (full
context, 1,024 tokens). The DistilBERT v2 tail-truncation strategy was
introduced in the first paper and retains the last 512 tokens rather
than the first, preserving the outcome-critical final turns of each
conversation. Hyperparameters and architectures are unchanged between
the pre-Path-B first-pass training run
and the post-cleanup retrain on the post-Path-B dataset, supporting
direct dataset-state comparison between the two runs (see
Section~\ref{sec:sensitivity}). Table~\ref{tab:hyperparams} summarizes
the configurations.

\begin{table}[h]
\caption{Classifier Configurations (post-cleanup retrain;
unchanged from pre-Path-B baseline).}
\label{tab:hyperparams}
\centering
\footnotesize
\begin{tabular}{lp{5.0cm}}
\toprule
\textbf{Model} & \textbf{Configuration} \\
\midrule
XGBoost + TF-IDF      & sklearn defaults; TF-IDF unigrams + bigrams,
                         5{,}000 max features, sublinear TF; plus 28
                         engineered features from the first paper's
                         feature set \\
\addlinespace
DistilBERT v2 (tail)  & \texttt{distilbert-base-uncased} (66M params);
                         tail truncation to last 512 tokens;
                         batch size 16, learning rate
                         $2 \times 10^{-5}$, 8 epochs, balanced class
                         weights; best-by-val-F1m model selection
                         (best epoch 4) \\
\addlinespace
Longformer (full)     & \texttt{allenai/longformer-base-4096}
                         (148M params); full-context input
                         (max 1{,}024 tokens, no truncation);
                         batch size 4, learning rate
                         $2 \times 10^{-5}$, 8 epochs, balanced class
                         weights; best-by-val-F1m model selection
                         (best epoch 5) \\
\bottomrule
\end{tabular}
\end{table}

\subsection{Results}

Table~\ref{tab:results} presents the complete 3-class test results for
all three models across both datasets. The key finding is preserved
from the pre-Path-B first-pass run and strengthened by the post-cleanup retrain:
Longformer surpasses XGBoost on all three metrics, reversing the ranking
from the first paper.

\begin{table}[h]
\caption{Complete 3-Class Test Results: First Paper vs.\ COVA-X
(post-cleanup retrain, post-VK-Path-B). Best results in bold.}
\label{tab:results}
\centering
\begin{tabular}{lrrr}
\toprule
\textbf{Model} & \textbf{Acc.} & \textbf{F1-m} & \textbf{F1-w} \\
\midrule
\multicolumn{4}{l}{\textit{COVA-X (10,985 conversations, this paper)}} \\
\textbf{Longformer (full)}      & \textbf{0.7971} & \textbf{0.7786} & \textbf{0.7956} \\
XGBoost + TF-IDF                & 0.7843          & 0.7563          & 0.7825          \\
DistilBERT v2 (tail)            & 0.7753          & 0.7632          & 0.7730          \\
\midrule
\multicolumn{4}{l}{\textit{First Paper Baseline (3,201 conversations)}} \\
XGBoost + TF-IDF                & 0.7250          & 0.6910          & 0.7180          \\
DistilBERT v2 (tail)            & 0.6980          & 0.6740          & 0.6950          \\
Longformer (full)               & 0.6980          & 0.6670          & 0.7010          \\
\midrule
\multicolumn{4}{l}{\textit{Improvement (COVA-X vs.\ First Paper, same model)}} \\
Longformer                      & $+9.91$pp       & $+11.16$pp      & $+9.46$pp       \\
XGBoost + TF-IDF                & $+5.93$pp       & $+6.53$pp       & $+6.45$pp       \\
DistilBERT v2                   & $+7.73$pp       & $+8.92$pp       & $+7.80$pp       \\
\bottomrule
\end{tabular}
\end{table}

Longformer achieves 0.7971 accuracy and 0.7786 macro F1, exceeding
XGBoost (0.7843, 0.7563) and DistilBERT v2 (0.7753, 0.7632) on every
reported metric. The first paper's hypothesis that transformer
architectures would benefit from larger conversational training corpora
is directly confirmed.

\subsection{Per-Class Analysis}

Table~\ref{tab:perclass} reports test-set precision, recall, and F1
by class for all three models. We report all three classifiers rather
than Longformer alone because the cross-architecture per-class comparison
constitutes part of the cross-architecture consistency finding discussed
in Section~\ref{sec:cleanup_consistency}.

\begin{table}[h]
\caption{Per-Class Test Performance on COVA-X Test Set
(n=1,099, post-cleanup retrain). Best F1 per class in bold.}
\label{tab:perclass}
\centering
\footnotesize
\begin{tabular}{llrrrr}
\toprule
\textbf{Model} & \textbf{Class} & \textbf{n} & \textbf{Prec.} & \textbf{Rec.} & \textbf{F1} \\
\midrule
                       & complied  & 166 & 0.706 & 0.651 & 0.677 \\
XGBoost + TF-IDF       & partial   & 429 & 0.739 & 0.718 & 0.728 \\
                       & rejected  & 504 & 0.843 & 0.885 & 0.864 \\
\midrule
                       & complied  & 166 & 0.738 & 0.747 & 0.743 \\
DistilBERT v2 (tail)   & partial   & 429 & 0.752 & 0.678 & 0.713 \\
                       & rejected  & 504 & 0.803 & 0.867 & 0.834 \\
\midrule
                       & complied  & 166 & 0.696 & 0.771 & \textbf{0.731} \\
Longformer (full)      & partial   & 429 & 0.790 & 0.702 & \textbf{0.743} \\
                       & rejected  & 504 & 0.837 & 0.887 & \textbf{0.861} \\
\bottomrule
\end{tabular}
\end{table}

Longformer achieves the best F1 on every class. The transformer advantage
is most pronounced on the minority class \textit{complied} (Longformer
0.731 vs.\ XGBoost 0.677 F1, a 5.4 point gap), consistent with prior
findings that contextual encoders handle class imbalance better than
bag-of-words approaches at sufficient training-data scale. XGBoost's
weakness on \textit{complied} stems from low recall (0.651) despite
moderate precision (0.706) --- it under-predicts the minority class.
DistilBERT v2 is intermediate on \textit{complied} but underperforms
both other classifiers on \textit{partial} (F1 0.713), the largest
non-rejected class, reflecting its 512-token truncation limit
discarding mid-conversation context that Longformer's full-context
encoding preserves.

\subsection{Confusion Matrix Analysis}

Table~\ref{tab:confusion} presents the test-set confusion matrices for
all three classifiers on the post-cleanup retrain (n=1,099). Reporting all
three matrices side-by-side supports the cross-architecture consistency
analysis in Section~\ref{sec:cleanup_consistency}: the error structure
should be similar across architectures if the residual classifier
errors reflect dataset-level boundary ambiguity rather than
architecture-specific failure modes.

\begin{table}[h]
\caption{Confusion Matrices on COVA-X Test Set (n=1,099,
post-cleanup retrain). Rows = true class, columns = predicted class.}
\label{tab:confusion}
\centering
\footnotesize
\begin{tabular}{l|rrr}
\multicolumn{4}{l}{\textbf{XGBoost + TF-IDF}} \\
\toprule
\textbf{True \textbackslash{} Pred.} & \textbf{compl.} & \textbf{partial} & \textbf{rejected} \\
\midrule
complied  & 108 & 53  & 5   \\
partial   & 43  & 308 & 78  \\
rejected  & 2   & 56  & 446 \\
\bottomrule
\end{tabular}
\vspace{4pt}

\begin{tabular}{l|rrr}
\multicolumn{4}{l}{\textbf{DistilBERT v2 (tail)}} \\
\toprule
\textbf{True \textbackslash{} Pred.} & \textbf{compl.} & \textbf{partial} & \textbf{rejected} \\
\midrule
complied  & 124 & 35  & 7   \\
partial   & 38  & 291 & 100 \\
rejected  & 6   & 61  & 437 \\
\bottomrule
\end{tabular}
\vspace{4pt}

\begin{tabular}{l|rrr}
\multicolumn{4}{l}{\textbf{Longformer (full)}} \\
\toprule
\textbf{True \textbackslash{} Pred.} & \textbf{compl.} & \textbf{partial} & \textbf{rejected} \\
\midrule
complied  & 128 & 31  & 7   \\
partial   & 48  & 301 & 80  \\
rejected  & 8   & 49  & 447 \\
\bottomrule
\end{tabular}
\end{table}

The error structure is broadly consistent across all three classifiers:
the largest off-diagonal cells are in the \textit{partial} row,
distributed between \textit{complied} predictions (38--48 conversations)
and \textit{rejected} predictions (78--100 conversations). This boundary
confusion is expected: victims who engage extensively before rejecting
produce dialogue that is lexically similar to those who engage without
fully committing. The cross-architecture similarity of this error
pattern argues that the boundary cases are genuinely ambiguous at the
data level, not artifacts of any one architecture.

Two architecture-specific patterns are also worth noting. First,
DistilBERT v2 has the highest \textit{partial $\rightarrow$ rejected}
misclassification rate (100 of 429, or 23.3\%), consistent with its
512-token truncation discarding mid-conversation engagement signals
that distinguish partial compliance from outright rejection. Second,
XGBoost has the highest \textit{complied $\rightarrow$ partial}
misclassification rate (53 of 166, or 31.9\%), reflecting the
bag-of-words representation's difficulty distinguishing fully-committed
from partially-committed victims when both produce similar lexical
patterns.

\subsection{Pre/Post-Path-B Sensitivity}
\label{sec:sensitivity}

To isolate the contribution of the dataset cleanup pipeline
(Section~\ref{sec:quality}) from architectural and hyperparameter
choices, we report results at two pinned dataset states: a pre-Path-B
first-pass run on the initial COVA-X dataset, and a post-cleanup
retrain on the post-Path-B dataset reported above. Hyperparameters and training procedures are unchanged
between the two runs. The only varying input is dataset state: the
cleanup pipeline replaced the COVA-X expansion's initial-batch
(two-role architecture) virtual-kidnapping conversations with the
three-role Path B regenerations described in
Section~\ref{sec:quality}; the 1,330 virtual-kidnapping conversations
in the dataset are now exclusively three-role Path B output. The
cleanup pass also applied seven outcome-label corrections to
Carol-profile conversations.

\begin{table}[h]
\caption{Pre vs.\ Post-Path-B Classifier Performance (held-out test set,
n=1{,}099). Same hyperparameters; varying input is dataset state.}
\label{tab:sensitivity}
\centering
\small
\setlength{\tabcolsep}{4pt}
\begin{tabular}{lcccc}
\toprule
\textbf{Model} & \textbf{Pre acc.} & \textbf{Post acc.} & \textbf{$\Delta$ acc.} & \textbf{$\Delta$ F1-m} \\
\midrule
XGBoost + TF-IDF\footnotemark[1]  & 0.7764 & 0.7843 & $+0.79$pp & $+0.0198$ \\
DistilBERT v2 (tail)              & 0.7625 & 0.7753 & $+1.28$pp & $+0.0220$ \\
Longformer (full)                 & 0.7880 & 0.7971 & $+0.91$pp & $+0.0083$ \\
\bottomrule
\end{tabular}
\end{table}
\footnotetext[1]{The XGBoost pre-Path-B baseline shown here was not
separately serialized as a standalone JSON; its values are preserved
in the comparison sections of the April 4, 2026 DistilBERT v2 and
Longformer training output files, which independently logged
identical XGBoost reference metrics
(0.7764\,/\,0.7365\,/\,0.7713). The 5-class XGBoost run from April 2
is preserved with the same records for reference. The DistilBERT v2
and Longformer pre-Path-B baselines are anchored to JSON
training-output files preserved in the same records.}

\begin{figure}[h]
\centering
\begin{tikzpicture}
\begin{axis}[
    width=8.5cm,
    height=4cm,
    ybar=2pt,
    bar width=10pt,
    enlarge x limits=0.18,
    ymin=0, ymax=2.5,
    ylabel={Improvement (pp)},
    symbolic x coords={XGBoost, DistilBERT v2, Longformer},
    xtick=data,
    nodes near coords,
    nodes near coords align={vertical},
    legend style={
        at={(0.5,-0.20)},
        anchor=north,
        legend columns=2,
        font=\footnotesize,
        /tikz/every even column/.append style={column sep=0.5cm}
    },
    label style={font=\footnotesize},
    tick label style={font=\footnotesize},
    every node near coord/.append style={font=\tiny}
]
\addplot[fill=blue!30, draw=blue!60] coordinates {
    (XGBoost, 0.79)
    (DistilBERT v2, 1.28)
    (Longformer, 0.91)
};
\addplot[fill=orange!30, draw=orange!70] coordinates {
    (XGBoost, 1.98)
    (DistilBERT v2, 2.20)
    (Longformer, 0.83)
};
\legend{$\Delta$ accuracy (pp), $\Delta$ macro F1 ($\times 10^{2}$)}
\end{axis}
\end{tikzpicture}
\caption{Cross-architecture classifier improvement between the
pre-Path-B first-pass run and the post-cleanup retrain
(post-Path-B). All three architectures improve on both metrics.
Macro F1 improvement scaled by $10^{2}$ for visual comparability with
accuracy in \% point terms.}
\label{fig:sensitivity}
\end{figure}
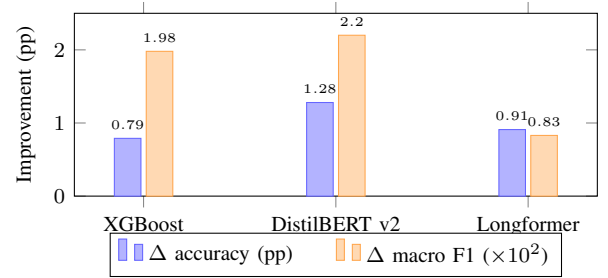

All three architectures improve on the post-Path-B dataset on both
accuracy and macro F1. Longformer remains the best-performing model
at both dataset states; the cross-architecture ranking is preserved
across the cleanup. Figure~\ref{fig:sensitivity} visualizes the
cross-architecture improvement pattern.

We discuss the implications of this consistency in
Section~\ref{sec:cleanup_consistency}.

\subsection{Five-Class XGBoost Comparison}
\label{sec:fiveclass}

The classifier results above use the 3-class collapse standardized
since the first paper (Section~\ref{sec:experimentation}). For
completeness, we report the corresponding 5-class XGBoost result on
the pre-Path-B baseline run, which was the first sklearn
training run on the COVA-X dataset and used the original 5-class
label set (\texttt{successful\_scam}, \texttt{partial\_compliance},
\texttt{verification\_attempt}, \texttt{scam\_detected},
\texttt{quick\_rejection}). On the same n=1{,}099 test set with
seed 42 stratification, XGBoost + TF-IDF achieves accuracy 0.7498,
macro F1 0.7745, weighted F1 0.7489. The macro F1 is higher in the
5-class formulation than 3-class, which is an artifact of two
small classes (\texttt{quick\_rejection} n=73,
\texttt{scam\_detected} n=97) being highly identifiable through
distinct lexical patterns; macro-averaging over the 5 classes
upweights these strong-performing minority classes. Substantively
the 5-class results are consistent with the 3-class headline:
XGBoost performs strongly on \texttt{quick\_rejection} (precision
0.784, recall 0.945) and \texttt{scam\_detected} (precision 0.914,
recall 0.763), but the boundary between
\texttt{successful\_scam}/\texttt{partial\_compliance} and
\texttt{verification\_attempt} remains the dominant error mode.
The 3-class formulation collapses these into the more
interpretable \textit{rejected} category and is retained as the
primary reporting basis for direct comparison to the first paper.

\section{Discussion}
\label{sec:discussion}

\subsection{Hypothesis Confirmation}

The first paper's central hypothesis---that transformer models would
benefit from larger training corpora---is directly confirmed by the
COVA-X results. Longformer's improvement from 0.667 to 0.7786 macro F1
($+11.16$ \% points) substantially exceeds XGBoost's improvement
from 0.691 to 0.7563 ($+6.53$ \% points). With 8,787 training
conversations, Longformer's full-context architecture is able to exploit
conversation-level signals that TF-IDF representations compress into
bag-of-words statistics. The cross-paper improvement gap (Longformer
$+11.16$pp vs.\ XGBoost $+6.53$pp on macro F1) is the primary
quantitative evidence supporting this paper's central claim.

\subsection{Why Longformer Surpasses XGBoost}
\textbf{Firstly,full-context encoding without truncation:} Longformer's
4,096-token capacity accommodates all conversations without information
loss (maximum observed token count: 874, well within the 1,024-token
training limit). In contrast, DistilBERT's 512-token limit still forces
truncation for approximately 20\% of sampled conversations even with
the tail strategy, discarding outcome-critical signal located in
mid-conversation turns.

\textbf{Secondly, sufficient data for fine-tuning:} 
Longformer exhibited overfitting by epoch 4 (training loss 0.18 vs.\
validation loss 2.30) in \cite{cova1}. On COVA-X, Longformer's best epoch is 5, and
the validation loss remains below 1.0 at that point, indicating
substantially better generalization from the larger training set.

\subsection{Cross-Architecture Cleanup-Effect Consistency}
\label{sec:cleanup_consistency}

The Pre/Post-Path-B Sensitivity results in Table~\ref{tab:sensitivity}
provide a methodological observation distinct from the headline
hypothesis confirmation. All three classifier architectures improve
on the post-Path-B dataset relative to the pre-Path-B first-pass run,
on both accuracy and macro F1. The cleanup intervention---reduction
in stage-direction artifacts, dialogue-collapse instances, and
repetition loops in virtual-kidnapping conversations, plus seven
high-confidence outcome relabels---produces measurable signal
recoverable across architectural families that operate on
fundamentally different representational substrates (bag-of-words
TF-IDF, transformer embeddings with tail truncation, transformer
embeddings with full context).

This consistency is informative because it argues against the cleanup
operating on architecture-specific noise. If the post-Path-B
improvements were concentrated in tokens visible only to one model
(for example, only to the tail-truncation window of DistilBERT v2),
we would expect asymmetric improvement patterns across the three
architectures. Instead, all three improve, suggesting the cleanup
affected genuine label-relevant signal accessible at multiple levels
of representational granularity.

We note, however, that the improvement magnitudes are not uniform
across architectures. DistilBERT v2 shows the largest absolute change
on both accuracy ($+1.28$pp) and macro F1 ($+0.0220$). XGBoost shows
a smaller swing on accuracy ($+0.79$pp) and a comparable swing on
macro F1 ($+0.0198$). Longformer shows the smallest swing on macro F1
($+0.0083$). One plausible reading is that DistilBERT v2's larger
swing reflects the proximity of cleanup to its truncation window: the
cleanup primarily affected end-of-conversation tokens (stage
directions, dialogue-collapse text, repetition loops in late VK
turns), which are precisely the tokens the tail-truncation strategy
preserves. Cleaner end-of-conversation tokens produce an outsized
DistilBERT improvement.\footnote{It is worth flagging
that XGBoost's small swing despite substantial cleanup could be
read either as capability ceiling (the model has extracted what
bag-of-words can extract from the lexical signal, and additional
cleanup does not unlock additional features) or as a robustness
property (the model was less sensitive to the artifact noise removed
by cleanup in the first place). We do not adjudicate between these
readings; the relevant claim for this paper is the cross-architecture
\emph{direction} of improvement, not its precise mechanism per
architecture.} Longformer's smaller swing is consistent with its
architecture being less sensitive to artifact concentration in any
particular conversational region: with full context available, it
already extracted most of the available signal at the pre-Path-B
state, so cleanup adds proportionally less.

We treat this as a methodological observation rather than a causal
claim. Cleanly establishing causality between cleanup-by-token-region
and per-architecture improvement magnitude would require a controlled
ablation isolating cleanup effects to specific token windows, proposed as future work in Section~\ref{sec:future_work}.

The cross-architecture consistency claim is supported by direct
JSON-anchored evidence for two of three classifiers. The
DistilBERT v2 and Longformer pre-Path-B baselines are preserved as
pre-cleanup training-output JSON files, paired against the
post-cleanup retrain JSONs.

\subsection{Quality Pipeline as Methodology}

The 3.9\% relabeling correction rate in COVA-X versus 49.8\% in the
first dataset represents a $12.7\times$ improvement in label
self-consistency. This improvement was not a single fix but the
compounded result of five independent pipeline improvements: profile
pinning for Carol and George, the ABSOLUTE OUTPUT RULE injection,
the glob$\rightarrow$rglob bug fix in the relabeling script, expanded
victim/attacker profiles providing more behavioral variety, and
revised attacker prompt templates eliminating cross-type contamination.

 The Pre/Post-Path-B Sensitivity results
(Section~\ref{sec:cleanup_consistency}) provide additional methodological
evidence beyond the label-cleanliness story. The cleanup
pipeline---a separate intervention from the audit-and-relabel pass
that yielded the 3.9\% rate---produced cross-architecture
classifier improvements indicating that the cleanup affected
classifier-recoverable signal. We argue that documenting these
improvements quantitatively, with before/after deltas at each
pipeline stage and at multiple classifier architectures, is itself
a methodological contribution. Synthetic dataset construction is an
iterative process, and the growing pains encountered are informative
for future researchers.

\subsection{Limitations}

First, the dataset remains English-only
and US-centric. Second, all models still treat conversations as
complete documents; real-time detection would require incremental
classification (see Section~\ref{sec:future_work}). Third, legacy
grandparent-opener contamination persists in government impersonation,
medicare, and virtual kidnapping files from pre-fix pipeline runs;
specific counts (gov\_imp 357, medicare 348, VK 355) are documented
in the companion data sheet and preserved as legacy contamination
rather than corrected post-hoc. Fourth, we document below a set of
generation-model capability limits identified through iterative
prompt engineering on the virtual kidnapping scam type. These
findings are specific to Qwen 2.5 14B and motivate the
model-comparison study proposed in Section~\ref{sec:future_work}.

\subsubsection{Structural Tag Adoption Failure}
Virtual kidnapping is uniquely structured among the eight scam types
in COVA-X: it involves three parties (attacker, victim, and a
supposedly-kidnapped hostage whose voice is played by AI-cloned audio
during the call). Our generation pipeline is architecturally two-role,
producing alternating attacker and victim turns. We initially attempted
to embed hostage voice inside attacker turns using structural tags
such as \texttt{[HOSTAGE VOICE]:}, with the tag adopted by the attacker
model during generation.

Across four iterations of prompt engineering (v6 through v9 of the VK
attacker prompt), we progressively intensified enforcement of this tag
rule through explicit prohibitions, forbidden-format examples,
repositioned rule priority, and recovery instructions. Despite these
reinforcements, Qwen 2.5 14B produced the target tag in 0\% of attacker
turns across 30 test conversations. The model instead defaulted to a
pretraining-learned bare-colon pattern (e.g., \texttt{: ``Mom! Help!''})
in every attacker turn regardless of the prompt structure. This suggests
that the attacker-hostage voice embedding is not a prompt-engineering
problem at the 14B-parameter scale under this emotional context.

\subsubsection{Salience Bias in Prompt Engineering}
During Path A iterations, we observed an instructive anti-pattern.
Version 9 of the attacker prompt explicitly enumerated forbidden
bare-colon variants (e.g., \texttt{: ``Dad!''}, \texttt{: ``Please!''})
as negative examples, intending to discourage their production.
Counterintuitively, the bare-colon leakage rate \emph{increased} from
approximately 15\% of attacker turns under v8 to approximately 40\%
under v9. We interpret this as salience-bias in instruction-tuned
LLMs: adding examples of prohibited patterns increases their
attention-weight presence in the prompt context, making the model
more likely to reproduce them rather than suppress them. This finding
is methodologically useful for practitioners attempting prompt-level
constraint enforcement on similar generation tasks.

\subsubsection{Architectural Workaround and Its Residual Limits}
To address the structural tag adoption failure, we introduced a
three-role generation architecture treating the hostage as a distinct
agent during generation, with a dedicated system prompt and its own
turn (Path B). At JSON-write time, hostage turns are folded backward
into the preceding attacker turn with the \texttt{[HOSTAGE VOICE]:}
tag added by the generator itself, preserving two-role format for
downstream classifier training. This architectural change achieved
100\% structural consistency by construction and eliminated the
dialogue-collapse failure mode observed in earlier batches.
Section~\ref{sec:quality} reports the three-point evidence chain
documenting Path B's effect on VK-specific artifact rates
(67.1\% pre-Path-B $\rightarrow$ 46.7\% post-Path-B raw $\rightarrow$
46.5\% post-Path-B processed flag rate).

The three-role architecture did not, however, address several
additional Qwen 2.5 14B behaviors:

\begin{itemize}
  \item \textit{Repetition loops under emotional pressure.} Attacker
  turns frequently converge on near-verbatim repetition by turn 10 in
  high-emotional-pressure scenarios, consistent with context-window
  degradation rather than role confusion.

  \item \textit{Profile compliance breakdown.} The highly-skeptical
  and informed victim profiles (targeting \textit{quick\_rejection}
  and \textit{scam\_detected} outcomes, respectively) showed
  compliance breakdown under VK emotional pressure. The model treats
  personality-profile directives as soft guidance rather than hard
  constraints in emotionally charged contexts. This behavior is
  observed consistently in VK and partially in government
  impersonation; other scam types respect victim profiles.

  \item \textit{Pretraining distribution bias on address terms.}
  Regardless of the configured \texttt{hostage\_address\_term}
  parameter (Mom/Dad/Grandma/Grandpa/Honey), Qwen 2.5 14B defaults
  to ``Mom!'' address patterns in approximately 70\% of grandchild
  and spouse relationships, reflecting a strong pretraining prior
  toward child-to-mother kidnapping-scenario audio.

  \item \textit{Cross-conversation name drift.} The most frequent
  dataset names (e.g., Jessica, Ashley) occasionally bleed across
  distinct victim profiles, producing concatenated-name artifacts
  (e.g., ``JessicaEmily'') indicating the model's output is
  influenced by recent training exposure despite explicit
  name-identity rules in the system prompt.

  \item \textit{Conversation termination override.} Government impersonation prompts include a CONVERSATION ENDING rule, added during the initial expansion run and subsequently strengthened, instructing termination after victim refusal or success. The model still overrides this rule in about 22.3\% of conversations, producing farewell-and-compliance loops at turns 15--19. Because stronger enforcement risks the salience-bias regression noted above, these loops are treated as a documented capability limit rather than further prompt-engineered away.
\end{itemize}
These five behaviors, observed independently across distinct
prompt-directive types (structural tag adoption, victim-profile
compliance, address-term parameterization, name-identity rules,
and conversation-termination directives), constitute a consistent
pattern: Qwen 2.5 14B honors prompt directives in short conversations
and low-pressure contexts, but overrides them under sustained
emotional or contextual pressure as conversation length increases.
We treat this as a capability ceiling rather than a prompt-engineering
deficiency, motivating the model-comparison study proposed in
Section~\ref{sec:future_work}. The five-instance pattern provides a
well-defined test target for future work: can a comparison model
obey structural tags, sustain profile specifications under emotional
pressure, parameterize address terms, maintain name identity, and
honor conversation-termination directives across the same test
conditions?

Residual artifact rates are addressed post-generation through a
bare-colon stripping filter and non-Latin-script filter in the
cleaning pipeline, with the post-cleanup VK flag rate of 46.5\%
documented in Section~\ref{sec:quality}. Mitigation through
architectural and post-processing changes is partial; future work
using alternative or larger generation models is the expected path
to further reduction.

\section{Conclusion and Future Work}
\label{sec:future_work}

This paper presented COVA-X, expanding the COVA conversational smishing dataset from 3,201 to 10,985 synthetic multi-turn conversations. Retraining all classifiers on the expanded corpus shows that Longformer now outperforms XGBoost + TF-IDF, achieving 79.71\% accuracy and 0.7786 macro F1 compared with 78.43\% and 0.7563 for XGBoost. This supports the hypothesis from our prior work that transformer models require larger conversational corpora to realize their contextual advantages. Across model families, the strongest gains were observed for transformer-based models, indicating that contextual representations become more effective as dataset scale and quality improve.

COVA-X also contributes a synthetic-data quality lifecycle analysis. Pipeline refinements reduced relabeling corrections from 49.8\% to 3.9\%, while the virtual-kidnapping three-role architecture reduced artifact flag rates from 67.1\% to 46.7\%, with post-processing yielding a final rate of 46.5\%. Per-scam-type outcome analysis shows mechanism-consistent patterns, including higher \texttt{successful\_scam} rates for virtual kidnapping, higher \texttt{verification\_attempt} rates for grandparent scams, and substantial \texttt{quick\_rejection} rates for medicare and bank impersonation. Pre/post-cleanup sensitivity analysis further shows that cleanup improved label-relevant signal across multiple classifier architectures.

Future work will extend COVA-X to audio-domain voice-phishing detection using synthetic text-to-speech, conduct controlled ablations to separate dataset-scale effects from pipeline-quality effects, and evaluate alternative generation models and grammar-constrained decoding to address observed generation failures. Additional directions include real-time classification from partial transcripts, cross-linguistic expansion, real-transcript validation, per-scam-type classifier analysis, and early-stopping studies for transformer fine-tuning. \textbf{The COVA-X dataset, accompanying data sheet, classifier baselines, and evaluation scripts will be made available for research use following acceptance.}


\end{document}